\documentclass[conference]{IEEEtran}
\IEEEoverridecommandlockouts
\usepackage{cite}
\usepackage{pifont}       
\usepackage{bbding}       
\usepackage{fontawesome}  
\usepackage{arydshln}
\usepackage{amsmath,amssymb,amsfonts}
\usepackage{algorithmic}
\usepackage{graphicx}
\usepackage{textcomp}
\usepackage{color}

\usepackage[normalem]{ulem}
\usepackage{multirow}
\useunder{\uline}{\ul}{}
\def\BibTeX{{\rm B\kern-.05em{\sc i\kern-.025em b}\kern-.08em
    T\kern-.1667em\lower.7ex\hbox{E}\kern-.125emX}}
\usepackage{booktabs}

\usepackage{url}

\usepackage[numbers,sort&compress]{natbib}
\usepackage{hyperref}

\begin{document}

\title{

Triple GNNs: Introducing Syntactic and Semantic Information for Conversational Aspect-Based Quadruple Sentiment Analysis

\thanks{* Corresponding author: Bingnan Ma.}
}

\author{
    \IEEEauthorblockN{
        Binbin Li\textsuperscript{1}, 
        Yuqing Li\textsuperscript{1,2}, 
        Siyu Jia\textsuperscript{1}, 
        Bingnan Ma\textsuperscript{3*},
        Yu Ding\textsuperscript{1}
    }
    \IEEEauthorblockN{
        Zisen Qi\textsuperscript{1},
        Xingbang Tan\textsuperscript{1}, 
        Menghan Guo\textsuperscript{1}, 
        Shenghui Liu\textsuperscript{1}
    }
    \IEEEauthorblockA{\textsuperscript{1}Institute of Information Engineering, Chinese Academy of Sciences}
    \IEEEauthorblockA{\textsuperscript{2}School of Cyber Security, University of Chinese Academy of Sciences}
    \IEEEauthorblockA{\textsuperscript{3}The National Computer Network Emergency Response Technical Team/Coordination Center of China}
     \IEEEauthorblockA{\{libinbin, liyuqing, jiasiyu, dingyu, qizisen, tanxingbang, guomenghan, liushenghui\}@iie.ac.cn, mbn@cert.org.cn}
}

\maketitle

\begin{abstract}

Conversational Aspect-Based Sentiment Analysis (DiaASQ) aims to detect quadruples \{target, aspect, opinion, sentiment polarity\} from given dialogues.
In DiaASQ, elements constituting these quadruples are not necessarily confined to individual sentences but may span across multiple utterances within a dialogue. This necessitates a dual focus on both the syntactic information of individual utterances and the semantic interaction among them.
However, previous studies have primarily focused on coarse-grained relationships between utterances, thus overlooking the potential benefits of detailed intra-utterance syntactic information and the granularity of inter-utterance relationships.
This paper introduces the Triple GNNs network to enhance DiaAsQ. It employs a Graph Convolutional Network (GCN) for modeling syntactic dependencies within utterances and a Dual Graph Attention Network (DualGATs) to construct interactions between utterances.
Experiments on two standard datasets reveal that our model significantly outperforms state-of-the-art baselines. The code is available at \url{https://github.com/nlperi2b/Triple-GNNs-}.

\end{abstract}

\begin{IEEEkeywords}
Conversational aspect-based sentiment analysis, Graph Neural Network, Discourse structure
\end{IEEEkeywords}

\section{Introduction}
As conversational applications\cite{luo2022critical} become more influential in real-world scenarios and sentiment analysis\cite{yu2023syngen,zhang2022survey} tasks advance, the need for dialogue-level sentiment quadruple analysis (DiaASQ)\cite{DBLP:conf/acl/Li0LWZWLLLCJ23} has emerged. DiaASQ aims to extract all quadruples $\bf{\{t,a,o,p\}}$ from dialogues, encompassing target ($\bf{t}$), aspect ($\bf{a}$), opinion ($\bf{o}$), and corresponding sentiment polarity ($\bf{p}$).
While Aspect-Based Sentiment Analysis (ABSA) focuses on fine-grained sentiment analysis of individual sentences, DiaASQ extends this analysis to extract sentiment quadruples from multi-turn dialogues.
As illustrated in Fig.~\ref{fig:data_example}, dialogues with multiple speakers discussing various products represent a multi-turn dialogue. In this dialogue, participants comment on aspects (e.g., `photography') of specific targets (e.g., `Apple'), expressing opinions (e.g., `so bad'). The sentiment polarity (e.g., `negative') is deduced from the combination of the target and opinion.
Notably, elements of a quadruple can be distributed within an utterance or across different utterances, as exemplified in Fig.~\ref{fig:data_example}(b).
Therefore, DiaASQ faces two challenges:
\begin{figure}[!htbp]
\centerline{\includegraphics[width=0.7\linewidth]{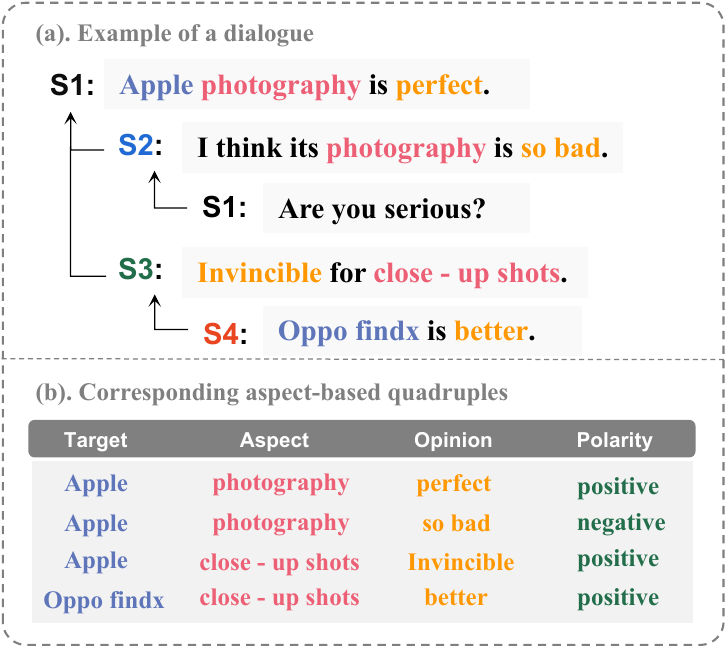}}
\caption{An illustration of a tree-like dialogue and its corresponding sentiment quadruples. Here, S1 through S4 represent the speakers of the utterances. Arrows indicate reply relations between the utterances.}
\label{fig:data_example}
\end{figure}
\begin{itemize}
    \item \textbf{Intra-utterance Quadruple Extraction}: 
Similar to the difficulties faced in ABSA tasks\cite{chen2021semantic,zhang2022boundary}, it is crucial to improve the extraction of quadruples from sentences. This is especially important when dealing with complex situations like overlapping quadruples, including one-to-many and many-to-one relationships.
    \item \textbf{Cross-utterance Quadruple Extraction}:  Unique to multi-turn dialogues, this challenge emphasizes the need for accurately extracting quadruples that span across different utterances, highlighting the importance of explicitly capturing the semantic and contextual information within dialogue.
\end{itemize}

Previous work\cite{DBLP:conf/acl/Li0LWZWLLLCJ23} initially introduced this benchmark and developed multi-view interactions (speaker, reply, and thread) to capture dialogue-specific features. However, it primarily focused on exploring coarse-grained relationships between utterances, neglecting the rich syntactic structure information within individual utterances and the potential effectiveness of finer-grained semantic interactions in the dialogue.

In this paper, we propose a Triple GNNs model that considers both the syntactic information within utterances and the semantic relationships between them.
Specifically, we introduce the \textit{intra-syntax GCN} module to capture syntactic information by modeling syntactic dependency tree and part-of-speech of utterances, thereby enhancing the extraction of intra-utterance quadruples. Simultaneously, we develop the \textit{inter-semantic GATs} module to capture the semantic and contextual interactions between utterances by utilizing speaker information and discourse structure. Finally, we integrate syntactic information and semantic information to improve DiaASQ.
Our contributions can be summarized as follows:
\begin{itemize}
    \item We introduce a novel Triple GNNs network to integrate intra-utterance syntactic information and inter-utterance semantic information, enhancing the DiaASQ task. Consequently, both intra-utterance quadruples and cross-utterance quadruples can be more effectively extracted.
    \item We exploit a GCN to model syntax through dependency trees, initializing word embeddings with Part-Of-Speech information. Additionally, we utilize two GATs to capture semantic information, including speaker-aware context and discourse structure within the dialogue.
    \item We conduct extensive experiments on two public datasets and our method achieves state-of-the-art performance.
\end{itemize}

\section{Related Works}\label{related_work}
\subsection{Aspect-based sentiment analysis}
Aspect-Based Sentiment Analysis (ABSA)\cite{yu2023syngen,zhang2022survey,jiang2023semantically} is dedicated to the fine-grained analysis of reviews. ABSA research revolves around four key sentiment elements: aspect (a), opinion (o), category (c), and sentiment polarity (p). Based on studies related to one or more of these elements, ABSA has given rise to various sub-tasks, ranging from single ABSA tasks to compound ABSA tasks.
The single-ABSA task aims to predict just one sentiment element. Examples of such tasks include Aspect Term Extraction (ATE), Opinion Term Extraction (OTE), Aspect Category Detection (ACD), and Aspect Sentiment Classification(ASC).

While single-ABSA tasks provide insights into specific sentiment elements, they often do not comprehensively address user requirements. To address this limitation, compound-ABSA tasks\cite{huang2023prompting,zhang2024adaptive,he2019interactive} have been developed to simultaneously extract multiple sentiment elements. These tasks can be categorized into three types: pair, triplet, and quadruple extractions.
Pair extraction\cite{he2019interactive} focuses on extracting two sentiment elements, encompassing Aspect-Opinion extraction, end-to-end ABSA, and Aspect Category Sentiment Analysis. Triplet extraction\cite{eberts2019span,jiang2023semantically} delves deeper by analyzing three distinct elements within a sentence. Although these methods have achieved success, there remains a gap in fully extracting all four sentiment elements.
To bridge this gap, the emerging area of quadruple prediction\cite{cai2021aspect,zhang2021aspect} aims to detect all four sentiment elements \{a, o, c, p\} from reviews.
These subtasks operate on individual sentences, which implies that elements of the quadruple only exist within a single sentence. 
\subsection{Emotion Recognition in Conversation}
Emotion Recognition in Conversation \cite{majumder2019dialoguernn,ghosal2019dialoguegcn,zhang2023dualgats,li2021past}focuses on identifying the emotion of each utterance within a given dialogue. The core challenge in this task is to consider the comprehensive contextual history and speaker-specific details of the target utterance in order to recognize its emotion.
Early research\cite{majumder2019dialoguernn} employed recurrent neural networks (RNNs) to grasp the contextual representation by discerning sequential relationships. Subsequent works\cite{ghosal2019dialoguegcn}, adopted a graph-based approach, treating each utterance as a node and used graph neural networks (GNNs)\cite{kipf2016semi} to frame ERC as a node-classification task.
Recent studies\cite{li2022neutral,zhang2023dualgats} have incorporated auxiliary knowledge into ERC to augment conversational context information, including dialogue structure, VAD (Valence, Arousal, Dominance), and personality traits.
Comparatively, both ERC and DiaASQ tasks operate on dialogues, requiring an explicit exploration of contextual information and pertinent dialogue-specific characteristics. 
However, ERC is primarily a classification task, seeking only to categorize emotions from a predefined set. In contrast, DiaASQ aims to extract a comprehensive range of sentiment elements from dialogues.
\section{Method}\label{Section_Method}
\begin{figure*}[htbp!]
\centerline{\includegraphics[width=0.68\linewidth]{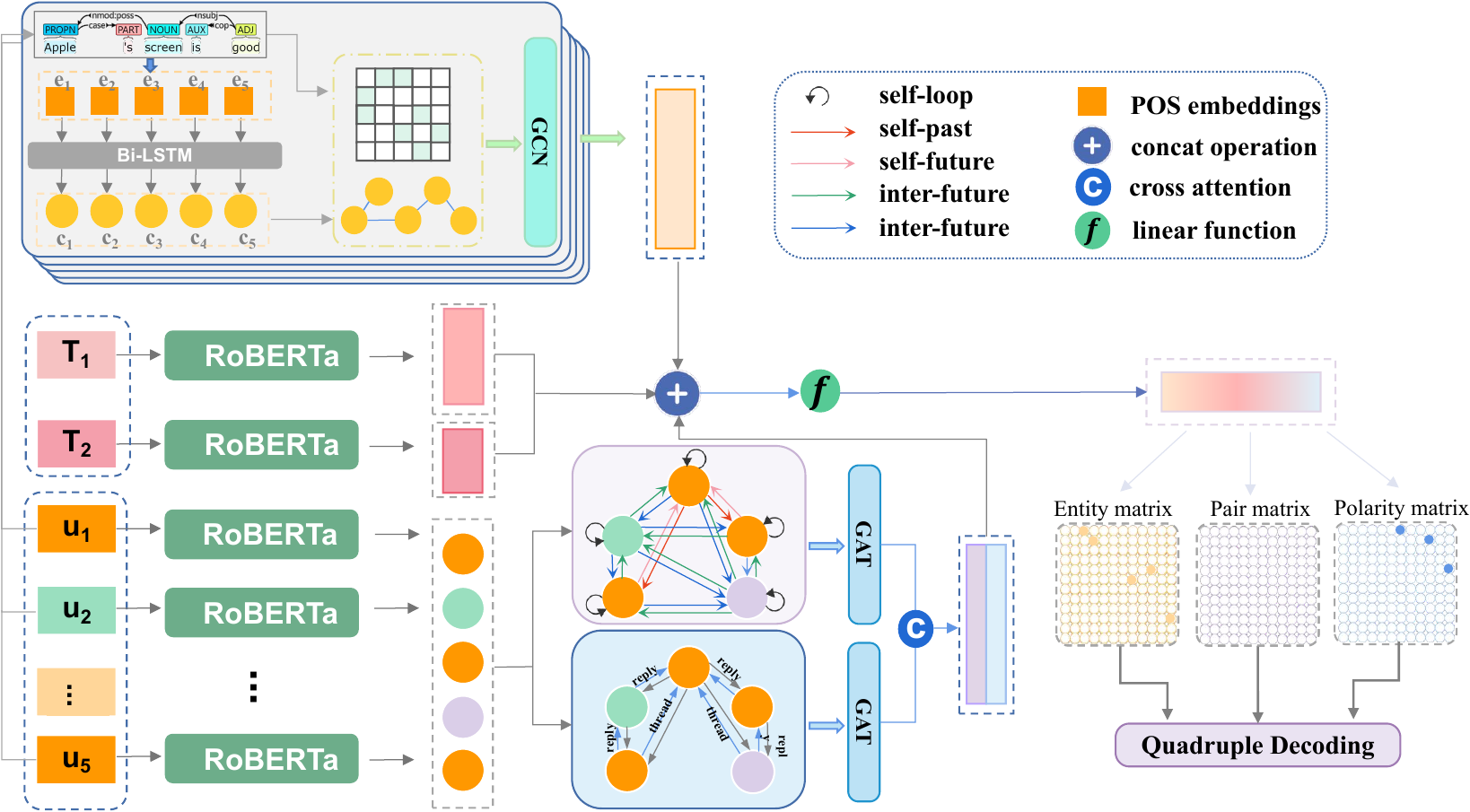}}
\caption{The architecture of our proposed Triple GNNs, which includes four main components: the encoder, a syntactic graph convolution network (Syn-GCN), a dual graph attention network, which includes a speaker-aware GAT (Spk-GAT) and a discourse structure-aware GAT (Str-GAT), and decoding matrices.}
\label{fig:model}
\end{figure*}

In this section, we introduce our Triple GNNs model. First, we define the task formulation, and then we introduce our proposed modules, which include the intra-GCN module and inter-GATs module. The architecture is shown in Fig.~\ref{fig:model}.
\subsection{Task Formulation}
Given a dialogue \(D = \{(u_i, s_i, r_i) | i = 1, ..., N\}\), the dialogue consists of \(N\) utterances \(\{u_i\}_{i=1}^N\). \(\{s_i\}_{i=1}^N\) and \(\{r_i\}_{i=1}^N\) represent the speaker and response information, respectively. To elaborate further, each utterance \(u_i\) is spoken by speaker \(s_i\) and is a response to a preceding utterance \(u_{r_i}\)
The primary objective of DiaASQ is to predict all sentiment quadruples \(Q = \{(t, a, o, p)\}\) within a dialogue. Each quadruple comprises the following elements: target (\(t\)), aspect (\(a\)), opinion (\(o\)), and polarity (\(p\)). Polarity is categorized into three predefined classes: \{positive, negative, other\}.

Following previous study\cite{DBLP:conf/acl/Li0LWZWLLLCJ23}, we decompose the task into three distinct subtasks: detection of entity boundaries, entity pairs, and sentiment polarity. We use \{tgt, asp, opi\} to denote the entities corresponding to the target, aspect, and opinion, respectively. Additionally, we use \{t2t, h2h\} to represent the relationships between these entities. Finally, the sentiment polarities are indicated using \{pos, neg, other\}.

\begin{table}[!ht]
\fontsize{9}{11}\selectfont
\centering
\def\narrtablewidth{0.6 cm}
\def\maintablewidth{0.7 cm}
\def\widetablewidth{0.7 cm}
\def\maxtablewidth{0.7 cm}
\begin{tabular}{p{0.43 cm} p{0.68 cm}
    p{\narrtablewidth} p{\maintablewidth} p{\widetablewidth} 
    p{\narrtablewidth} p{\maintablewidth} p{\widetablewidth} 
    p{\maxtablewidth} p{\maintablewidth} p{\maxtablewidth} 
    p{\narrtablewidth} p{\maintablewidth} p{\widetablewidth}}
\hline
 & & \multicolumn{3}{c}{\bf ZH-dataset}  & \multicolumn{3}{c}{\bf EN-dataset} \\
\cmidrule(r){3-5}\cmidrule(r){6-8}
  & & \#Train. & \#Valid. & \#Test. & \#Train. &\#Valid. &  \#Test. \\
\hline
\multirow{1}{*}{\bf Dia.}  &
 & 800 & 100 & 100   & 800 & 100 & 100\\
\multirow{1}{*}{\bf Utt.}  &
& 5,947 & 748 & 757 & 5,947 & 748 & 757 \\
\multirow{1}{*}{\bf Spk.}  &
&3,986 & 502 & 503&3,986 & 502 & 503\\
\multirow{1}{*}{\bf Quads.}  &
& 4,607 &577 & 558 & 4,414 & 555 & 545\\
\hline
\end{tabular}
\caption{Statistics for datasets. `Dia', `Utt', `Spk', and `Quads' denote `dialogue', `utterance', `speaker', and `quadruples'.}
\label{table_corpus_statistics}
\end{table}
\subsection{Base Encoding}
\subsubsection{Single Utterance Encoding}
We employ RoBERTa\cite{liu2019roberta} as our encoder. Due to the input length constraints, encoding an entire dialogue directly is not feasible. Therefore, we individually encode each utterance to obtain its representation, which will be utilized for encoding the discourse structure.
For a given utterance \(u_i = [w^i_1, w^i_2, \ldots, w^i_{n}]\) comprising \(n\) words, its individual encoded result is as follows:
\begin{equation}
\small
{h_i} = \text{RoBERTa}(\tilde{u_i})
\label{eq:1}
\end{equation}
\begin{equation}
\small
  \tilde{u_i} = [[CLS], w^i_1, w^i_2, ..., w^i_{n}, [SEP]]
  \label{eq:2}
\end{equation}
where ${h_i}\in \mathbb{R}^{n\times d}$ and $d$ denotes the hidden dimension. The tokens $[CLS]$ and $[SEP]$ are treated as special tokens to indicate the beginning and end of an utterance, respectively.
\subsubsection{Thread Encoding}
With the additional conversational information present in the DiaASQ task, which enables the dialogue to form a tree-like structure based on reply information, each branch of the dialogue tree is considered as a separate \textit{thread}. Our approach expands the context window, extending the encoding target from a single utterance to a whole thread. This accommodates input length and enhances the capture of richer contextual information. The input thread is defined as a composition of several related utterances:
\begin{equation}
\small
\textbf{T} = \bigcup_{t=1}^{|T|}T_t = \bigcup_{t=1}^{|T|}\{u_1, u_{k_t}, ..., u_{k+\gamma}\}
\end{equation}
Where \(k_t\) is the position of the second node in the \(t\)-th branch. \(\gamma\) denotes the position of the leaf node of that branch. 

These threads are then encoded individually to obtain thread-level representations \(G = \{{g_1}, ..., {g_{|T|}}\}\). The encoding mechanism is analogous to that described in Equations~\ref{eq:1} and~\ref{eq:2}.


\subsection{Triple GNNs network}
\subsubsection{Intra-Syntax GCN}
The main challenge in quadruple extraction lies in accurately extracting entity elements and matching relationships. Word-level linguistic features provided by Part-Of-Speech(POS) tags offer potential clues for the boundary recognition of entity spans\cite{chen2021semantic,wu2021learn,jiang2023semantically}. Additionally, syntactic dependency trees can guide the matching of relationships between entities.  Therefore, we utilize POS tags and syntactic dependency trees obtained from an external third-party parser for intra-utterance quadruple extraction in dialogues.
Specifically, for each utterance \(u_i\), we first obtain a list of POS tags \(\{p^i\}_{n}\). These POS tags are then passed through an embedding layer, yielding learnable continuous embedding vectors \({\mathbf{e}^i_p}\). To capture the context of the sentence, we employ a Bi-LSTM to process \({\mathbf{e}^i_p}\), obtaining POS representation enriched with contextual information: 
\begin{equation}
\small
    \mathbf{e}^i_p = \text{Embedding}([p^i_1, p^i_2, ..., p^i_{n}])
\end{equation}
\begin{equation}
\small
    \mathbf{c}^i = \text{Bi-LSTM}(\mathbf{e}^i_p)
\end{equation}
where \(\mathbf{e}^i_p\in \mathbb{R}^{n\times d_e}\) and \(\mathbf{c}^i\in \mathbb{R}^{n\times d_l}\). \(d_e\) and \(d_l\)  denote the dimensionality of embeddings and the hidden state, respectively.

To integrate POS information with syntactic dependency information, we formulate a syntactic graph, \(\mathcal{G}^{syn}=(\mathbf{V}^{syn},\mathbf{E}^{syn})\), where \(\mathbf{V}^{syn}\) represents the set of words and \(\mathbf{E}^{syn}\) denotes the edges between them, representing syntactic dependencies. 
Firstly, we define an adjacency matrix \(A^{syn}=\{a_{ij}\}_{n\times n} \) in \(\mathbf{E}^{syn}\), where \(a_{ij}\) indicates the dependency relationship between words \(w_i\) and \(w_j\). We then use the hidden state vectors \(\mathbf{c}^i\) from the BiLSTM as the initial node representations in the syntactic graph. The syntactic graph representation for utterance \(u_i\) is given as \(H_i^{syn}=\{h^{syn}_1,...,h^{syn}_n\}\).
 We denote \(h_i^{syn(l)}\) as the representation of word \(w_i\) at the \(l\)-th layer:
\begin{equation}
\small
    h_i^{syn(l)} = \text{ReLU}\left(\sum_{j=1}^n a_{ij} W^l h_i^{syn(l-1)}+b^l\right)
\end{equation}
Where \(W^l\) is the weight matrix and \(b^l\) is the bias term. The initial representation \(h_i^{syn(0)}\) is initialized by \(c_i\). The syntactic hidden representation for the entire conversation is denoted as: $\mathbf{H}^{syn}=\sum_{i=1}^NH_i^{syn},\mathbf{H}^{syn}\in\mathbb{R}^{N\times n \times d_e}$.
\subsubsection{Inter-Semantic GATs}
The effectiveness of speaker and contextual cues in addressing dialogue-related works is well-established\cite{li2022neutral,zhang2023dualgats}.
Observing similar trends, we find that in DiaASQ, these elements are crucial for connecting information across utterances. Inspired by \cite{zhang2023dualgats}, we implement Dual GATs to discern complex dependency structures and semantic interrelations among utterances in a dialogue. Our approach is distinguished by leveraging inherent speaker and response information within the dialogue dataset for graph construction, eliminating the reliance on external data and thus avoiding additional noise. Furthermore, we develop dependency relationship types specifically designed for quadruple extraction in dialogue sentiment analysis.
As  depicted in Fig.~\ref{fig:model}, 
this module comprises two main components: the speaker-aware GAT and the discourse structure-aware GAT.

\textbf{Speaker-Aware GAT.}
We first define the speaker dependency graph as \(\mathcal{G}^{spk}=(\bold{V}^{spk},\bold{E}^{spk})\). The node set \(\bold{V}^{spk}\) represents each utterance \(u_i\) in the dialogue. The representation of these nodes is initialized with the encoded representation \({h_i}\) of the utterance. 
The adjacency matrix \(\bold{E}^{spk}\) denotes the relationship between the speakers and nodes. Following the conventions of prior graph-based ERC works\cite{ghosal2019dialoguegcn,zhang2023dualgats}, we define five types of speaker dependencies: \textit{Self-Past, Self-Future, Inter-Past, Inter-Future}, and \textit{Self-Loop}. The `Self' implies that the compared pair of utterances originates from the same speaker, whereas `Inter' signifies utterances from different speakers.  `Past' and `Future' refers to the temporal order of the utterances, with the term `self-loop' being used to define the diagonal of the graph. Subsequently, we employ the GAT to capture the speaker-aware contextual information within the dialogue. This approach enables us to derive speaker-enriched utterance representations, denoted as \({h}_i^{spk}\). The computation process can be described as follows:
\begin{equation}
\small
\alpha_{ij} = \text{softmax}(\sigma(\mathbf{a}^T(\mathbf{W_i}h_i;\mathbf{W}_jh_j;\mathbf{E}_{ij})))
\label{eq:g1}
\end{equation}
\begin{equation}
\small
    {h}^{spk}_i = \sum_{j\in \mathcal{N}_i^{spk}}\alpha_{ij}W_k{h}_j
    \label{eq:g2}
\end{equation}
Where \(h_i\) and \(h_j\) represent the representations of nodes \(i\) and \(j\) respectively. \(\sigma\) is the LeakyReLU activation function.  \(\mathbf{a^T}, \mathbf{W_i}, \mathbf{W_i}\), and \(W_k\) are trainable parameters. \(\mathcal{N}_i^{spk}\) denotes the neighboring node set of node \(v_i^{spk}\). The representation of the entire conversation, represented as: \(\mathbf{H}^{spk}=\sum_{i=1}^N{h}^{spk}_i\).

\textbf{Structure-Aware GAT. }
The structure-aware GAT module is introduced to incorporate the discourse structure information among utterances in a conversation, focusing on their threading relationship, reply relationship, and temporal relationship. To initiate this process, we first craft a structure-aware graph as \( \mathcal{G}^{str} =(\mathbf{V}^{str},\mathbf{E}^{str}) \).
Similarly, utterances in a dialogue compose the set \( \mathbf{V}^{str} \), which is initialized by their corresponding representations \( \{{h_i} \}_n \). The weight of edge \( e_{ij}^{str} \) is determined by the type of relationship \( r_{ij} \) from the set \( R^{str} \). The set \( R^{str} \) comprises five types of structural dependencies: \textit{self-loop, reply-past, reply-future, thread-past,} and \textit{thread-future}, as depicted in Fig.~\ref{fig:model}.
We then utilize the GAT to obtain a structure-aware context for utterances. The computation process is similar to that in Equations.~\ref{eq:g1} and \ref{eq:g2}.
Finally, we engage an interaction module to integrate semantics. This process is formulated as follows:
\begin{equation}
\small
\begin{aligned}
A_1 & = \text{softmax}\left(\mathbf{H}^{spk} W_1 (\mathbf{H}^{str})^T\right) \\
A_2 & = \text{softmax}\left(\mathbf{H}^{str} W_2 (\mathbf{H}^{spk})^T\right) \\ 
\end{aligned}
\end{equation}
\begin{equation}
\small
\mathbf{H}^{spk'},\mathbf{H}^{str'}  = A_1\mathbf{H}^{spk}, A_2\mathbf{H}^{str}
\end{equation}
Where \( \mathbf{H}^{\lambda}\in \mathbb{R}^{N\times d} \), \( \lambda=\{spk',str'\} \), and both \( W_1 \) and \( W_2 \) denote trainable parameters.

\subsubsection{Dialogue Quadruple Prediction}
Having obtained the sentence-level syntactic dependency information feature \( \mathbf{H}^{syn} \in \mathbb{R}^{N \times n \times d} \), and the structural-semantic relationship features \( H^{spk'} \in \mathbb{R}^{N \times d} \) and \( H^{str'} \in \mathbb{R}^{N \times d} \) via our designated network, we aggregate this information with the encoded thread feature \( G \). This leads to an enhanced thread-level feature representation, \( \mathbf{H}^{Tri} \), which incorporates explicit intrasentential syntactic dependencies and part-of-speech (pos) information, as well as rich semantic dependencies between sentences. We then amalgamate these enhanced representations into dialog-level features \( \mathbf{H}^{\mathcal{D}} \) to facilitate the final extraction of the sentiment quadruple.
\begin{equation}
\small
    \mathbf{H}^{Tri} = \text{Linear} \left( \mathbf{H}^{syn} \oplus \mathcal{E}(\mathbf{H}^{spk'} \oplus \mathbf{H}^{str'}) \right)
\end{equation}
\begin{equation}
\small
 \mathbf{H}^{\mathcal{D}} = \sum_{t=1}^{|T|} \left[ \Sigma_{i \in T_t} \mathbf{H}_i^{Tri} \right] \oplus G 
\end{equation}
Where \( \oplus \) is the concatenation operation, \( \mathcal{E} \) represents tensor expansion, \( \mathbf{H}^{\mathcal{D}} \in \mathbb{R}^{N_{sum} \times d} \), and \( N_{sum} = \sum_{i=1}^N n_i \) denotes the total tokens in the dialog.

\subsubsection{Quadruple Decoding}
Initially, we employ the enhanced dialogue feature representation \( \mathbf{H}^{\mathcal{D}} \) to calculate scores for each token pair. Depending on the type of relationship, these features are processed through tag-specific MLP (Multi-Layer Perceptron) layers to derive the final representation \( v_i^{\tau} \). Finally, we decode all the quadruples by predefined rules.
\begin{equation}
\small
    v_i^{\tau} = \text{MLP}^{\tau}(\mathbf{h}_i^{\mathcal{D}})
\end{equation}
\begin{equation}
\small
    {p}^{\tau}_{ij} = \text{softmax}\left((v_i^{\tau})^T v_j^{\tau}\right)
\end{equation}
Where \( \tau \) $\in$\( \{tgt, asp, opi, h2h, t2t, pos, neg, other, \epsilon_{ent}\} \), with \( \epsilon_{ent} \) indicating a non-label relation.

\begin{table*}[h]
\fontsize{9.5}{11}\selectfont
\centering
      \resizebox{0.9\linewidth}{!}{$
  \begin{tabular}{l ccccc||ccccc}
  \hline
\multicolumn{1}{c}{\multirow{2}{*}{\small \text{\bf{Model}}}} & \multicolumn{5}{c||}{\text{\bf{ZH-dataset}}} & \multicolumn{5}{c}{\text{\bf{EN-dataset}}} 
  \\ 
  \cline{2-11}
  & {{Intra}}& {{Inter}}& {{micro-F1}} & {{iden-F1}} & {{avg-F1}} 
 &  {{Intra}}& {{Inter}}&{{micro-F1}} & {{iden-F1}}  & {{avg-F1}}  \\
  \hline
     CRF-Extract-Classify$^\dagger$ &/&/& 8.81 & 9.25  &9.03&/&/ & 11.59 & 12.80 & 11.71   \\ 
     SpERT$^\dagger$ &/&/& 13.00 & 14.19 &13.6 &/&/& 13.07 & 13.38 &13.23  \\
     ParaPhrase$^\dagger$ &/&/& 23.27 & 27.98 & 25.63 &/&/&24.54 & 26.76 &25.65   \\
    Span-ASTE$^\dagger$ &/&/& 27.42  & 30.85 &29.14&/&/ & 26.99  & 28.34 &27.67  \\
    DiaASQ$^\dagger$  &37.95&23.21& 34.94 & 37.51 &36.23  &37.65&15.76&  33.31 &  36.80 &35.10  \\
    \hline
    \textbf{Ours}  
    &\textbf{46.12}&\textbf{29.02}&\textbf{42.87}   & 	\textbf{45.43}   & 	\textbf{44.15} &\textbf{40.77}  &\textbf{27.91}
    &\textbf{38.32} &   \textbf{40.97} &\textbf{39.65}\\
  \hline
  \end{tabular}    
  $}
  \caption{We report the micro-F1 scores and identification F1\cite{barnes2021structured} scores for DiaASQ. `Intra' and `Inter' are micro f1 score for extraction of quadruples that exist in the same utterance and cross several different utterances. $\dagger$ means the results are copied from \cite{DBLP:conf/acl/Li0LWZWLLLCJ23}.
  }
  \label{tab:main}
\end{table*}

\subsubsection{Training Loss}
Our training goal is to minimize the following  objective function of each subtask and obtain the final total loss:
\begin{equation}
\small
    \ell_{\tau} = -\frac{1}{N_{sum}^2}\sum_{i=1}^{N_{sum}}\sum_{j=1}^{N_{sum}}\alpha^{\tau} y_{ij}^{\tau}\log(p_{ij}^{\tau})
\end{equation}
\begin{equation}
\small
    \mathcal{L} = \ell_{ent} + \ell_{pair} + \ell_{pol}
\end{equation}
where $\alpha^{\tau}$ is a tag-wise weighting vector.

\subsection{Experiment Settings}
\subsubsection{Dataset and Evaluation Metric}
We evaluate our framework on two benchmark datasets: the Chinese dataset(ZH) and the English dataset(EN), which are mobile phone field review conversations derived from the social platform Weibo\footnote {\url{https://weibo.com/}}. The statistics of these datasets are shown in Table.~\ref{table_corpus_statistics}. 

Consistent with prior research, we adopt the micro F1-score and identification-F1\cite{barnes2021structured} (iden-F1) as our evaluation metrics. While the micro F1-score evaluates the entire quadruple, the iden-F1 focuses solely on the triple \( (t,a,o) \) and does not account for sentiment polarity.

\subsection{Compared Methods}
To fairly evaluate the performance of the Triple GNNs model, we choose the following baselines for DiaASQ.

\begin{itemize}
    \item \textbf{ExtractClassify} \cite{cai2021aspect}: This approach divides quadruple extraction into two stages. The first stage involves the extraction of aspects and opinions, while the second stage focuses on predicting category sentiment based on the extracted aspect-opinion pairs.
    
    \item \textbf{SpERT} \cite{xu2021learning}: A range-based attention model that is specifically tailored for the joint extraction of entities and their relations.
    
    \item \textbf{Span-ASTE} \cite{eberts2019span}: This method introduces a span-based approach to address the limitations faced by end-to-end models, particularly in handling multi-word entities.
    
    \item \textbf{ParaPhrase} \cite{zhang2021aspect}: A generation-based approach that creates quadruples by paraphrasing original sentences, offering a novel perspective in quadruples generation.
    \item \textbf{DiaASQ} \cite{DBLP:conf/acl/Li0LWZWLLLCJ23}: The benchmark of DiaASQ. This method proposes a  multi-views interaction network, which aggregates threads, speakers, and replying information.
    
\end{itemize}

\subsection{Implementation Details}
For a fair comparison and consistent with previous research, we use Roberta-Large\cite{liu2019roberta} and Chinese-Roberta-wwm-base\cite{cui2021pre} as our base encoders for the English and Chinese datasets, respectively. To transform the input text into a syntax dependency tree, we utilize the open-source toolkit SPaCy\footnote{\url{https://spacy.io}}. We set the hidden dimension to 768 for Chinese and 1024 for English. The dropout rate is configured at 0.4, and the learning rate for BERT is set to 1e-5. Our training runs for 30 epochs using various random seeds, with a batch size of 1. All models are implemented using PyTorch, version 1.12.1. Our experiments are executed on a single NVIDIA Tesla P100 GPU equipped with 16GB of memory.

\section{Results and Discussions}
\subsection{Main Results}
The primary experimental results are presented in Table~\ref{tab:main}. Notably, our proposed method significantly outperforms the state-of-the-art on both the ZH and EN datasets, achieving higher average F1 scores by 7.92 and 4.55 points, respectively. Moreover, our approach consistently surpasses existing models in both the intra-utterance and inter-utterance quadruple extraction scenarios. This performance strongly supports the effectiveness of our Triple GNNs model in handling quadruples within dialogues. It excels not only in extracting quadruples from individual sentences but also in identifying and leveraging the relationships in dialogues for enhanced cross-utterance quadruple extraction. In summary, these findings conclusively demonstrate that our proposed method significantly improves the DiaASQ task.





\begin{table}[!ht]
\centering
\caption{Ablation Results on or Triple GNNs. `intra' represents F1 score of intra-utterance quadruples, and `inter' represents F1 score of cross-utterances quadruples.}
\label{ablation}
\renewcommand\arraystretch{1.2}
\begin{tabular}{lccc|ccc} 
\hline
\multirow{2}{*}{Model} & \multicolumn{3}{c}{ZH}  & \multicolumn{3}{c}{EN}   \\ 
\cline{2-7}
                 &intra &inter    & $F1$  &intra &inter\textbf{}  & $F1$                   \\ 
\hline

{Ours}  & {46.12} & \textbf{29.02} & \textbf{42.87} & \textbf{40.77}   &\textbf{27.91} &\textbf{38.32}                 \\
\hdashline

 \quad  $\bullet$ \bf{w/o intra-GCN}        & 45.68 & 26.79 & 41.81
 &39.20  &24.31   &36.3              \\

\hdashline
\quad $\bullet$ \bf{w/o inter-GATs}     & 46.06  & 26.34   & 42.14
 & 39.49 & 19.89 & 35.51  \\
 \quad \quad w/o speaker &\bf{46.86}&28.14&41.43 &39.40&24.72&36.54\\
\quad \quad w/o structure &44.47 &25.87 &40.94 &40.61&22.86&37.15\\
\hline

\end{tabular}
\label{tab:ablation}    
\end{table}

\subsection{Ablation Study and Analysis}
To evaluate the significance of each component in our Triple GNNs model, we carry out an ablation study, with the results depicted in Table~\ref{tab:ablation}. Initially, the elimination of the intra-GCN module leads to a marked decrease in overall performance, notably in intra-quadruple extraction. This indicates that syntactic information is crucial for understanding relationships between tokens within an utterance. Additionally, removing the inter-GAT module results in a consistent drop in performance. Furthermore, the omission of any element within the inter-GAT module, whether it’s the speaker-GAT or the discourse structure-GAT, adversely affects the overall model effectiveness.
Moreover, concurrently removing both the intra-GCN and inter-GATs modules leads to a significant reduction in performance. This further emphasizes the critical role of these two modules in our model.

\section{Conclusion}\label{section_conclusion}

In this study, we introduce a novel Triple GNNs network that concurrently integrates intra-utterance syntactic and inter-utterance semantic information for DiaASQ. By harnessing dialogue threads, we construct context windows. This approach allows the model to not only focus on the utterance itself but also ensures a thorough understanding of the surrounding dialogue context. Experimental results obtained from two datasets demonstrate the superiority of our Triple GNNs model. Additionally, through comprehensive evaluations and an ablation study, we have confirmed the effectiveness of our model and the significant impact of its components.

\bibliographystyle{IEEEtran}
\bibliography{reference}

\end{document}